\documentclass[acmtog]{acmart}
\AtBeginDocument{%
  }

\copyrightyear{2025}
\acmYear{2025}
\setcopyright{acmlicensed}\acmConference[SIGGRAPH Conference Papers '25]{Special
Interest Group on Computer Graphics and Interactive Techniques Conference
Conference Papers }{August 10--14, 2025}{Vancouver, BC, Canada}
\acmBooktitle{Special Interest Group on Computer Graphics and Interactive
Techniques Conference Conference Papers (SIGGRAPH Conference Papers '25), August
10--14, 2025, Vancouver, BC, Canada}
\acmDOI{10.1145/3721238.3730752}
\acmISBN{979-8-4007-1540-2/2025/08}
\author{Minghao Yin}
\orcid{0009-0001-4252-2179}
\affiliation{%
  \institution{The University of Hong Kong}
  \country{Hong Kong}
}
\author{Yukang Cao}
\orcid{0009-0001-0125-0015}
\affiliation{%
  \institution{Nanyang Technological
University}
  \country{Singapore}
}
\author{Songyou Peng}
\orcid{0009-0007-6085-8059}
\affiliation{%
  \institution{Google DeepMind}
  \country{USA}
}
\author{Kai Han}
\orcid{0000-0002-7995-9999}
\authornote{Corresponding author is Kai Han (kaihanx@hku.hk).}
\affiliation{%
  \institution{The University of Hong Kong}
  \country{Hong Kong}
}
\usepackage{lineno}
\usepackage{xcolor}
\usepackage{soul,xcolor}
\newcommand{\boldparagraph}[1]{\vspace{0.3em}\noindent{\bf #1.}}

\newcommand{\figref}[1]{Fig.~\ref{#1}}
\newcommand{\secref}[1]{Section~\ref{#1}}

\newcommand{\tabref}[1]{Table~\ref{#1}}

\begin{document}

\newcommand{\OM}{Splat4D}
\newcommand{\OMO}{Splat4D }

\sethlcolor{yellow}
\newcommand{\edit}[1]{\hl{#1}} % 高亮显示

\newcommand\yk[1]{{\color{violet}#1}}
\newcommand\q[1]{{\color{red}#1}}
\newcommand\kh[1]{{\color{magenta}{[KH: #1]}}}

\title{Splat4D: Diffusion-Enhanced 4D Gaussian Splatting for Temporally and Spatially Consistent Content Creation}

\begin{teaserfigure}
\centering
\includegraphics[width=16.9cm]{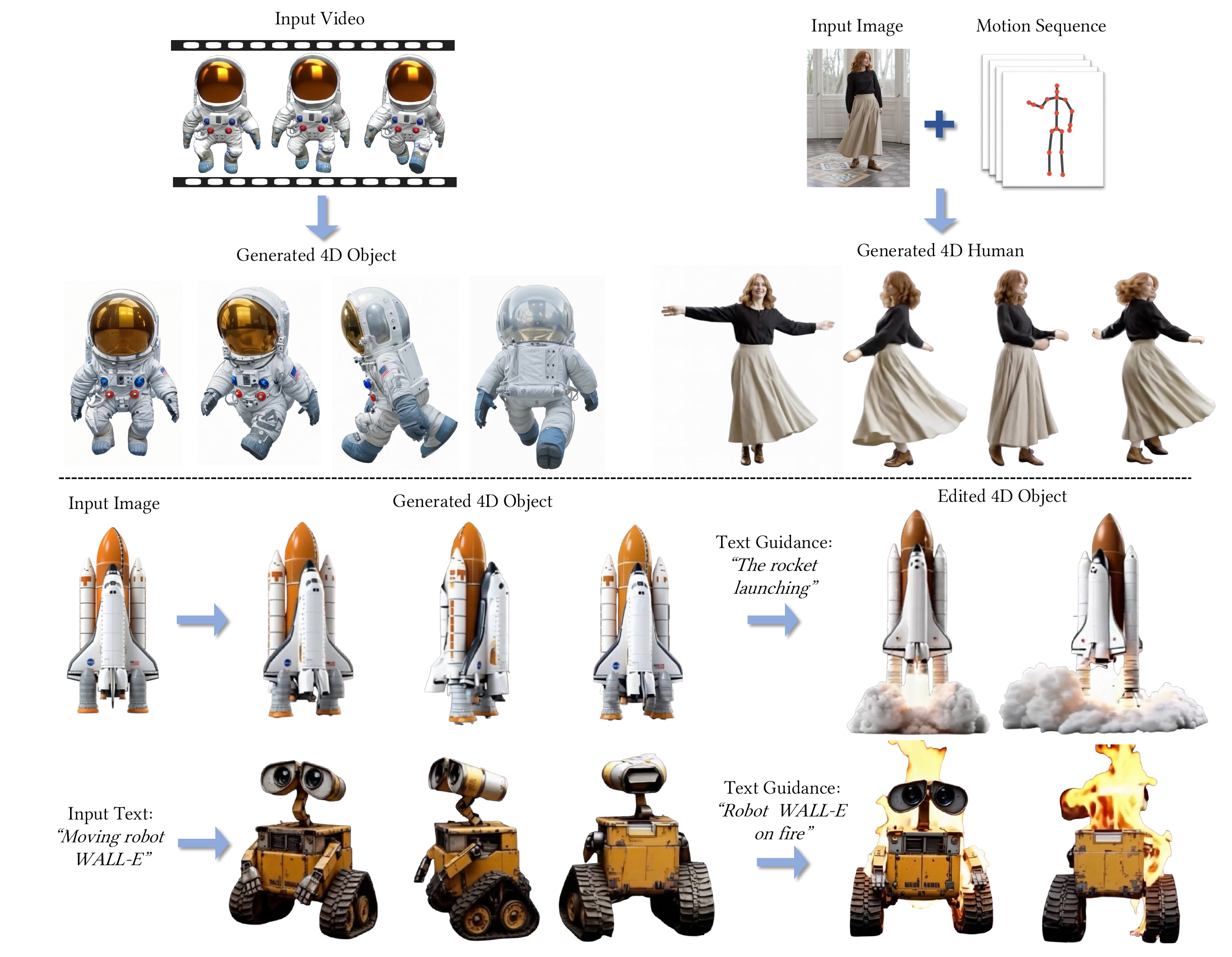}
%\caption{Our method enables versatile 4D content generation, creating dynamic 4D objects from text, image, or monocular video inputs and generating 4D human representations guided by image and motion sequence inputs. \kh{improve the caption, fix `` in the text in the figure, the left '' symbols are wrong.}}
\caption{\textbf{Splat4D.} Our method empowers a wide array of 4D content generation capabilities with high fidelity. Top-left: Generating the 4D representation from a monocular video; Top-right: 4D human generation guided by an image and motion sequence; Bottom-left: Generation of dynamic 4D objects from image or text inputs; Bottom-right: Text-guided 4D content editing, enabling detailed scene customization.}
%\centering
\label{fig:teaser}
\end{teaserfigure}
% \begin{abstract}
% We introduce a novel framework for generalizable 4D content creation from monocular video input, leveraging a 4D Gaussian splatting field for spatial and temporal consistency. Our method utilizes a diffusion model and image enhancer to produce multi-view coarse-stage image sequences, serving as the basis for initializing the 4D Gaussian representation. To enhance fidelity, we identify and mask inconsistent pixels within the image sequence, subsequently applying a video-denoising U-Net to reconstruct these regions and yield a fine-stage image sequence. This refined sequence further optimizes the 4D Gaussian splatting, enabling high-quality, dynamic content generation that maintains coherence across time and space. Our approach offers a promising solution for efficiently generating realistic, temporally stable 4D scenes from 2D video data.
% \end{abstract}

\begin{abstract}
Generating high-quality 4D content from monocular videos—for applications such as digital humans and AR/VR—poses challenges in ensuring temporal and spatial consistency, preserving intricate details, and incorporating user guidance effectively. To overcome these challenges, we introduce \OM, a novel framework enabling high-fidelity 4D content generation from a monocular video. 
\OMO achieves superior performance while maintaining faithful spatial-temporal coherence, by leveraging multi-view rendering, inconsistency identification, a video diffusion model, and an asymmetric U-Net for refinement. 
Through extensive evaluations on public benchmarks, \OMO  consistently demonstrates state-of-the-art performance across various metrics, underscoring the efficacy of our approach. Additionally, the versatility of \OMO is validated in various applications such as text/image conditioned 4D generation, 4D human generation, and text-guided content editing, producing coherent outcomes following user instructions. Project page: https://visual-ai.github.io/splat4d
\end{abstract}



  
\maketitle

\section{Introduction}

%\begin{figure*}
%\centering
%\includegraphics[width=\textwidth]{figs/teaser_new.pdf}
%\caption{Our method enables versatile 4D content generation, creating dynamic 4D objects from text, image, or monocular video inputs and generating 4D human representations guided by image and motion sequence inputs. \kh{improve the caption, fix `` in the text in the figure, the left '' symbols are wrong.}}
%\caption{\textbf{Splat4D.} Our method empowers a wide array of 4D content generation capabilities with high fidelity. Top-left: Generating the 4D representation from a monocular video; Top-right: 4D human generation guided by an image and motion sequence; Bottom-left: Generation of dynamic 4D objects from image or text inputs; Bottom-right: Text-guided 4D content editing, enabling detailed scene customization.}
%\centering
%\label{fig:teaser}
%\end{figure*}

The generation of 4D content---encompassing dynamic 3D objects---is integral to applications in digital humans, gaming, media, and AR/VR, where realistic motion and spatial-temporal consistency are essential. Unlike static 3D object generation~\cite{tang2024dreamgaussian, voleti2025sv3d, lin2023magic3d, wang2024prolificdreamer, raj2023dreambooth3d, melas2023realfusion, chen2023fantasia3d, tang2023make, qian2023magic123}, creating 4D content must capture an object's evolving appearance and motion within 3D space, which significantly increases the complexity. This challenge is especially pronounced when generating 4D content from a single monocular video, as it demands simultaneous inference of appearance and motion for unseen camera viewpoints. Moreover, the problem is inherently ill-posed, as multiple valid 4D interpretations can emerge from the same input. Consequently, representing 3D shape, texture, and motion in this high-dimensional space requires a substantial number of parameters, emphasizing the need for efficient modeling and computational strategies to address these demands.

Recent works~\cite{bahmani20244d, singer2023text, wang2025animatabledreamer, zheng2024unified, ren2023dreamgaussian4d, zhao2023animate124, cao2024avatargo} have explored 4D content generation through score-distillation sampling (SDS)~\cite{poole2023dreamfusion} loss with pre-trained diffusion models, producing dynamic scenes but often suffering from slow generation speeds and spatial-temporal inconsistencies. To address these limitations, follow-up approaches ~\cite{jiangconsistent4d, zeng2024stag4d, yin20234dgen} leverage 3D-aware diffusion models~\cite{shimvdream} to improve spatial consistency. 
Recently, with the development of video diffusion models~\cite{he2022latent, blattmann2023align, blattmann2023stable}, different techniques have been explored for fine-tuning these models to enable the generation of multi-view video sequences from single-view inputs. Leveraging this enhanced multi-view priors, methods such as 4Diffusion~\cite{zhang20244diffusion}, SV4D~\cite{xie2024sv4d}, and Diffusion4D~\cite{liang2024diffusiond} have been proposed to advance 4D generation and reconstruction more efficiently. Despite these advancements, significant challenges persist, including ensuring temporal and spatial consistency, accurately modeling complex human characteristics (e.g., loose clothing), and effectively integrating diverse input modalities such as images, text, and motion data.

To address these persistent challenges, we propose a novel framework for \emph{generalizable 4D content creation}, called \OM, which allows for high-quality 4D generation from a monocular video input, rendering versatile applications. 
First, we transform a monocular video into high-quality multi-view image sequences with a multi-view diffusion model and an image enhancer. 
% ~\cite{huang2024mv} and image enhancement~\cite{wang2021real}.
A pretrained large-scale feed-forward 3DGS model is then employed to obtain spatial and depth features across views, which are then further processed by Splatter Image to produce a comprehensive yet coarse Gaussian field.  
Next, we introduce a method for refining spatial and temporal coherence of the 4D Gaussian field representation through multi-view rendering, inconsistency identification, and a video diffusion model, resulting in a high-quality 4D representation with improved visual quality and stability.
Finally, to boost the realism of the generation, an asymmetric U-Net model is trained as the generalizable 3D Gaussian field predictor to produce accurate and detailed 3D Gaussians, improving overall quality.
% This sequence then initializes a coarse yet comprehensive 4D Gaussian splatting representation with a combination of a feedforward 3DGS model~\cite{tang2025lgm} and a splatter-image method~\cite{szymanowicz2024splatter}. 
% To refine the 4D coarse Gaussians and ensure consistency across temporal and spatial domains, we predict an uncertainty map to mask inconsistent regions, followed by a video denoising model~\cite{yu2024viewcrafter} to enhance rendering quality.
Our \OMO model is the first to incorporate an image enhancer for high-quality 4D generation, leading to substantial performance gains. However, directly applying the enhancer can cause multi-view inconsistencies and misalignments. To address this, we introduce uncertainty masking and asymmetric U-Net training to identify unreliable regions and adaptively refine the reconstruction. The video diffusion model then inpaints the masked areas, ensuring spatial-temporal consistency. These components are carefully integrated to complement each other, achieving results unattainable by any single module alone.
By seamlessly integrating these components, our \OMO framework can effectively generate high-quality spatial-temporal consistent 4D content at speed. Meanwhile, it can be applied for various of applications, such as text/image conditioned 4D generation, 4D human generation, and text-guided content editing (see~\figref{fig:teaser}).
% \kh{this part is not good, sounds like combining several existing methods.}

We thoroughly evaluate our framework on public benchmark, achieving the state-of-the-art results across the board on all metrics, validating the superior performance of our method. Moreover, we also showcase the results of applying our method for the applications of text/image conditioned 4D generation, 4D human generation, and text-guided content editing, demonstrating faithful and coherent results following the provided guidance.

\section{Related Work}
\boldparagraph{3D Generation} For 3D content generation, early works such as DreamFusion~\cite{poole2023dreamfusion} pioneer the use of Score Distillation Sampling (SDS) loss to distill priors from 2D diffusion models, optimizing 3D content from textual or image input. Subsequent efforts~\cite{wang2024prolificdreamer, tang2024dreamgaussian, yi2024gaussiandreamer, lisweetdreamer, weng2023consistent123, pan2024enhancing, chen2024text, sun2023dreamcraft3d, Sargent_2024_CVPR, liang2024luciddreamer, zhou2024dreampropeller, han2023headsculpt, cao2024dreamavatar, cao2023guide3d} address challenges like multi-view Janus artifacts, slow generation speed, and oversaturation by fine-tuning diffusion models for viewpoint control or directly generating multi-view images within a single diffusion pass. Methods like Zero123~\cite{liu2023zero} and SyncDreamer~\cite{liusyncdreamer} refine 2D diffusion models for multi-view consistency, while others, including Magic3D~\cite{lin2023magic3d} and Direct2.5~\cite{lu2024direct2}, adopt alternative 3D representations such as Instant-NGP~\cite{muller2022instant}, DMTet~\cite{muller2022instant}, or explicit mesh-based approaches~\cite{lu2024direct2} to improve runtime and fidelity.

%Another prominent direction focuses on generating dense multi-view images with strong 3D consistency, reconstructing 3D models based on these dense views. Approaches like 
DreamGaussian~\cite{tang2024dreamgaussian} introduces a point-based representation, utilizing 3D Gaussians for faster generation and superior quality compared to traditional Neural Radiance Fields (NeRF)~\cite{mildenhall2021nerf}. The feed-forward method LGM (Large Multi-View Gaussian Model)~\cite{tang2025lgm} efficiently represents scenes with multi-view Gaussian features and uses an asymmetric U-Net to process multi-view images. In our work, we draw inspiration from the ideas introduced in DreamGaussian~\cite{tang2024dreamgaussian}, which uses 3D Gaussians for faster, high-quality generation, and LGM~\cite{tang2025lgm}, which employs multi-view Gaussian features and asymmetric U-Nets to process multi-view images from single-view inputs.
% \songyou{what is the main relations/differences of our methods to those? It is quite unclear to me. Same for below.}

\boldparagraph{Video Diffusion Model} Recent video diffusion models have achieved impressive results in creating realistic motions and geometrically consistent sequences~\cite{ho2022video, voleti2022mcvd, blattmann2023align, blattmann2023stable, he2022latent, singermake, guoanimatediff}. Their strong generalization abilities stem from training on extensive image and video datasets, which are more accessible than large-scale 3D or 4D datasets. These models are increasingly utilized as foundational tools for tasks like multi-view synthesis and 3D content generation. For instance, frameworks like SV3D~\cite{voleti2025sv3d} adapt latent video diffusion models to produce consistent multi-view imagery, while approaches like AnimateDiff~\cite{guoanimatediff} enhance text-to-image models by incorporating motion modules to capture temporal dynamics. Similarly, SV4D~\cite{xie2024sv4d} employs video diffusion models to achieve both video generation and novel view synthesis. Leveraging these advancements, our approach extends pre-trained video generation models by introducing view attention mechanisms, aligning outputs for improved coherence in multi-view and 4D applications.

\boldparagraph{4D Generation} DreamGaussian4D~\cite{ren2023dreamgaussian4d} leverages a three-stage framework to generate high-quality 4D animations. It uses a modified version of Gaussian splatting combined with image-to-video diffusion for high-fidelity 3D static models that are deformed over time using a learned deformation field. DYST (Dynamic Scene Transformer)~\cite{seitzerdyst}  innovates further by decomposing monocular videos into distinct latent representations of scene content, per-view dynamics, and camera pose through co-training on real and synthetic data. GaussianFlow~\cite{gao2024gaussianflow} introduces Gaussian fields paired with optical flow constraints, further enhancing consistency in generated motion by aligning temporal transitions. 
AvatarGO~\cite{cao2024avatargo} proposes a correspondence-aware motion field that enables harmonious generation of 4D human–object interactions from text.

Building on the foundation of 3D-aware diffusion models~\cite{shimvdream}, recent methods such as Stag4D~\cite{zeng2024stag4d} and 4DGen~\cite{yin20234dgen} focus on enhancing spatial consistency in 4D generation. Meanwhile, Consistent4D~\cite{jiangconsistent4d} employs a video interpolation model to improve both temporal and spatial coherence. More recently, video diffusion models have been adopted for further 4D content enhancement. For instance, 4Diffusion~\cite{zhang20244diffusion} introduces a multi-view video diffusion model coupled with a 4D-aware SDS loss to optimize dynamic NeRF representations. Similarly, Diffusion4D~\cite{liang2024diffusiond} leverages a 4D-aware video diffusion framework combined with explicit 4D construction to synthesize 4D assets. For feed-forward generation, SV4D~\cite{xie2024sv4d} advances this line of work by utilizing a multi-view video synthesis model for efficient 4D optimization. Additionally, L4GM~\cite{ren2024l4gm} proposes a 4D interpolation model enabling fast feed-forward 4D generation from single-view video inputs.

However, current 4D generation methods often struggle to produce high-quality content that maintains both spatial and temporal consistency. Issues such as blurry textures, geometric distortions, and temporal flickering are still prevalent. To address these limitations, we introduce Splat4D, a novel approach that can generate high-quality, spatial-temporal consistent 4D content.

\section{Method}
Given a single-view image $I$ or a text prompt $y$ as input, \OMO facilitates the generation of a 4D dynamic scene. Specifically, our method captures both the spatial structure and the temporal evolution of the scene by decomposing it into multiple 3D Gaussian distributions. In the subsequent sections, we first illustrate the preliminaries that underpin our method in~\secref{sec:preliminaries}. We then delve into the core techniques of our method, including (1) coarse 4D Gaussian generation that leverages pre-trained diffusion priors in~\secref{sec:coarse}, (2) temporal and spatial refinement for improving consistency in~\secref{sec:refinement}, and (3) generalizable 3D Gaussian field predictor learning in~\secref{sec:unet-finetuning}.
An overview of our pipeline is shown in~\figref{fig:pipeline}.

% \yk{Given a \q{monocular video, $I=\left\{I_t\right\}_{t=1}^T$ where $T$ is the video length}, as input, \OMO \textcolor{red}{generates} a 4D Gaussian scene that }

\subsection{Preliminaries}
\label{sec:preliminaries}

\subsubsection{3D Gaussian Splatting} Different from NeRF~\citep{mildenhall2021nerf}, which relies on neural networks for novel view synthesis, 3D Gaussian Splatting (3DGS)~\citep{kerbl20233d} takes a different approach by directly optimizing the 3D position $\mathbf{x}$ and attributes of 3D Gaussians, such as opacity $\alpha$, anisotropic covariance, and spherical harmonic (SH) coefficients $\mathcal{SH}$~\citep{ramamoorthi2001efficient}. Each 3D Gaussian $G(\mathbf{x})$ is characterized by a 3D covariance matrix $\Sigma$ centered at its mean position $\mu$.
\begin{equation}
    G(\mathbf{x})=e^{-\frac{1}{2}(\mathbf{x}-\mu)^T \Sigma^{-1}(\mathbf{x}-\mu)}.
\end{equation}

For 3DGS, a tile-based rasterizer is utilized by dividing the screen into tiles, such as $16 \times 16$ pixels. For each tile it intersects, a Gaussian is instantiated and assigned a key encoding its depth in view space and the corresponding tile ID. Depth sorting is then applied to the Gaussians, enabling the rasterizer to efficiently resolve occlusions and overlapping structures. The final RGB color $\mathbf{C}$ is computed using a point-based $\alpha$-blending approach, which samples points along the ray at regular intervals.

\subsubsection{Large Multi-View Gaussian Model (LGM)} The LGM framework (e.g.~\cite{tang2025lgm}) transforms a single input image into a 3D Gaussian representation of the object through a systematic process. First, it employs MVDream~\cite{shimvdream}, a multi-view diffusion model, to generate multi-view consistent images from the input. MVDream synthesizes images from four viewpoints, ensuring they align geometrically and maintain visual consistency. These multi-view images are then processed by an asymmetric U-Net, the encoder extracts multi-scale spatial features from the images, while the decoder focuses on reconstructing these features into dense splatter images~\cite{szymanowicz2024splatter}. Splatter images encode the parameters for 3D Gaussians for each pixel, such as position, size, color, and opacity. These splatter images are then transformed into 3D Gaussian representations by backprojecting the pixel-wise Gaussian parameters into 3D space using the camera parameters.
% Each pixel's encoded information determines the Gaussian's information. 
We leverage the feedforward LGM for efficient but coarse scene representation generation.

\begin{figure*}[t]
\centering
\includegraphics[width=18cm]{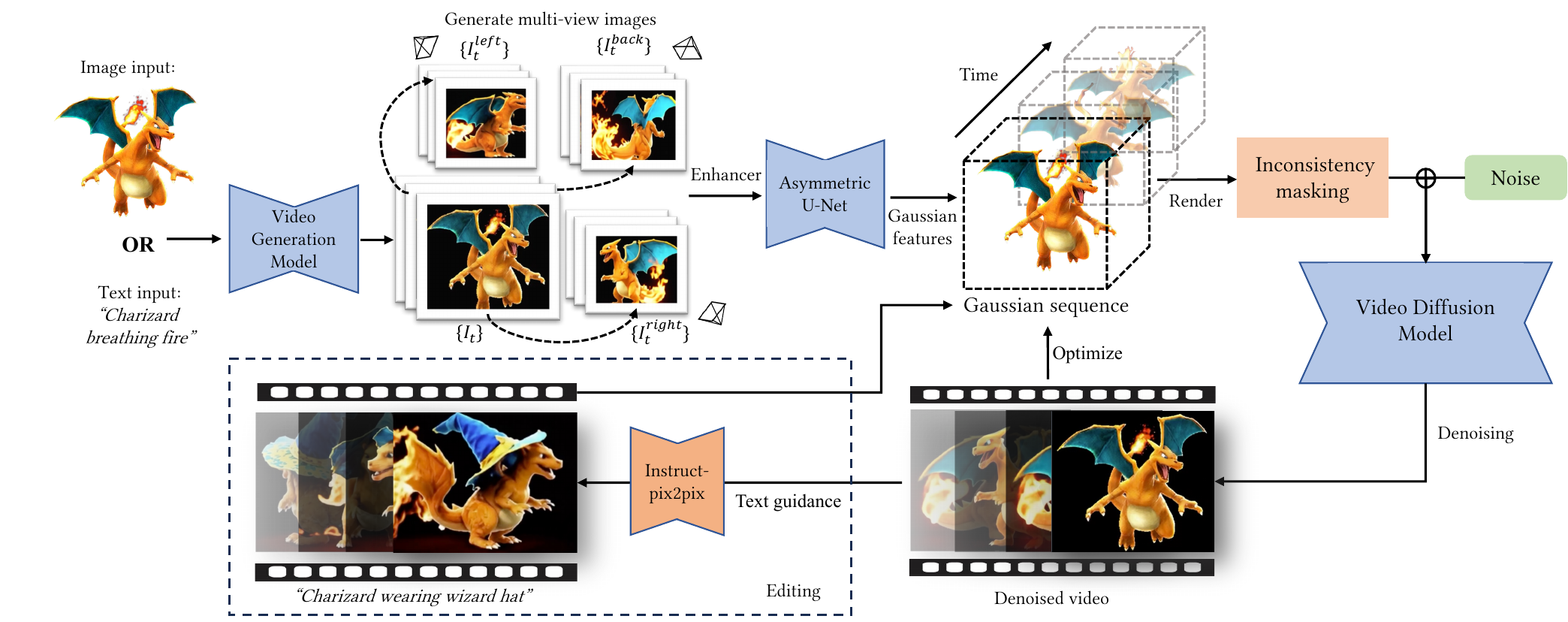}
\caption{\textbf{Overview of Splat4D}. Our method for 4D content generation begins with processing input data (text, image, or monocular video) to produce high-quality multi-view image sequences. These sequences are used to initialize a 4D Gaussian representation via an asymmetry U-Net and image splattering. Refinement steps include leveraging uncertainty masking and video denoising diffusion to ensure high fidelity and spatial-temporal consistency, culminating in versatile 4D content creation. The pipeline supports optional text-guided content editing, enabling dynamic modifications of the 4D output for enhanced flexibility and creative control.}
\label{fig:pipeline}
\centering
\end{figure*}

\subsection{Coarse 4D Gaussian Generation}
\label{sec:coarse}

Existing 4D generative methods~\cite{bahmani20244d, singer2023text, wang2025animatabledreamer, zheng2024unified, ren2023dreamgaussian4d, zhao2023animate124} typically employ pre-trained diffusion models~\cite{he2022latent} and score distillation sampling (SDS)~\cite{poole2023dreamfusion} to subsequently generate and animate 3D scenes from either text or image guidance. However, we observe two main limitations in such techniques: (1) it struggles to produce large, dynamic motions effectively, and (2) it requires significant training time to generate a single outcome. 
Drawing inspiration from the success of SV4D~\cite{xie2024sv4d}, our approach aims to overcome these limitations by first generating single-view videos and then distilling these videos into the 4D space for further refinement and animation.

\subsubsection{Multi-view Video Generation}

We utilize the video diffusion model~\cite{blattmann2023stable} to generate the image sequence. However, relying solely on a single-view video does not provide enough information for robust 4D modeling. This limitation stems from issues like depth ambiguity and the lack of side and back view information. To address this, we enhance the single-view video by using MV-Adapter~\cite{huang2024mv} to generate additional views, including the front, back, and sides, thereby enriching the model with more comprehensive rotational perspectives:
\begin{equation}
    \texttt{MV-Adapter}(I_t) \rightarrow \{I_t, I_t^{\text{left}}, I_t^{\text{right}}, I_t^{\text{back}} \},
\end{equation}
See Supplementary Material for evaluation for the choice of MV-Adapter over SV4D.

\subsubsection{Multi-view Image Enhancer}
Although MV-Adapter can robustly provide multi-view perspectives, their generated videos often lack fine-grained details and the high resolution required for realistic 4D content (see the figure in the supplementary material). This issue is expected as the input samples would always fall outside the training distribution of MV-Adapter. To address this, we propose to apply an image enhancer model~\cite{wang2018esrgan} \texttt{IE} to refine textures, edges, and details for each frame and each view.

% \begin{equation}
%     \{I_t, I_t^{\text{left}}, I_t^{\text{right}}, I_t^{\text{back}} \} = \texttt{IE}({I_t, I_t^{\text{left}}, I_t^{\text{right}}, I_t^{\text{back}}),
% \end{equation}

\subsubsection{4D Gaussian Reconstruction}
After generating a high-quality, multi-view image sequence, we proceed to construct a 4D Gaussian field. Following LGM~\cite{tang2025lgm}, we first input the multi-view image sequence into U-Net to encode key spatial and depth features across the views. The U-Net architecture is well-suited for this task because it can capture detailed structures at multiple resolutions through its encoder-decoder structure. The encoder captures feature maps at different scales, identifying essential textures and depth cues, while the decoder reconstructs these features into a cohesive representation. 

Once the U-Net has processed the multi-view sequence, we apply the Splatter Image~\cite{szymanowicz2024splatter} method to project these learned features into a continuous 4D Gaussian field. Specifically, Splatter Image maps each pixel from the feature maps into a series of localized Gaussian distributions in 3D space, with each Gaussian representing a small spatial region from the scene. To form the final temporally consistent Gaussian sequences, we design our network to separately reconstruct a 3D Gaussian field for each frame (time step). Specifically, we construct a stacked representation of multiple 3D Gaussians, represented as $\mathcal{G}(\mathcal{S}, t)=\left[\mathcal{X}_t, s_t, r_t, \sigma_t, \zeta_t\right]$, with position, scale, rotation, opacity and Spherical Harmonics (SH) at time $t$. This Gaussian field serves as a foundational structure, providing a spatially continuous and temporally stable representation that can be rendered from any angle. 
% The Gaussian field is represented as a set of 3D Gaussians at each time step.
%By integrating these Gaussians across the multi-view sequence for each time step, we create a coarse 4D field that embodies spatial information from various perspectives and maintains temporal coherence. }

% Considering LGM~\cite{tang2025lgm} only tackle the static scenario, we extend the concept of 3D Gaussians into a 4D space in our framework by constructing a stacked representation of multiple 3D Gaussians, represented as $\mathcal{G}(\mathcal{S}, t)=\left[\mathcal{X}_t, s_t, r_t, \sigma_t, \zeta_t\right]$, with position, scale, rotation, opacity and Spherical Harmonics (SH) at time $t$. This Gaussian field serves as a foundational structure, providing a spatially continuous and temporally stable representation that can be rendered from any angle. 
% This stage establishes the foundational structure necessary for subsequent refinement steps, in which we will address and enhance temporal and spatial consistency.

\subsection{Spatial-Temporal Consistency Refinement}
\label{sec:refinement}
Although multi-view video generation and image enhancement techniques can provide detailed 3D information necessary for constructing a 4D Gaussian scene, the resulting reconstruction still suffers from issues with temporal and spatial consistency. This happens because the MV-Adapter has difficulty maintaining consistent multi-view images. Additionally, since the MV-Adapter processes each frame independently, it further contributes to these inconsistencies in the model. 
To tackle this problem, we introduce a multi-step approach that involves two key techniques: \textit{inconsistency masking} and \textit{uncertainty-guided refinement}.

\subsubsection{Inconsistency Masking} We start by rendering a sequence of multi-view images $\{\mathcal{I}_t, \mathcal{I}_t^{\text{left}}, \mathcal{I}_t^{\text{right}}, \mathcal{I}_t^{\text{back}} | t \in [1, T]\}$ from the 4D Gaussian field $\mathcal{G}(\mathcal{S}, t)$, where $\mathcal{I}$ represents the rendered images. 
For each time step $t$, we then generate uncertainty maps~\cite{kulhanek2024wildgaussians} to detect regions with inconsistencies. We extract DINOv2~\cite{oquab2024dinov2} features from the rendered images and predict the pixel-wise uncertainty $\sigma$ using an uncertainty prediction network~\cite{kulhanek2024wildgaussians}. These uncertainty maps highlight areas that show significant variation or deviation between frames, which are often caused by issues like motion artifacts, occlusions, or perspective differences. By identifying these inconsistent areas, we create a mask that helps us focus on correcting the problematic regions while keeping the stable areas intact. The uncertainty mask is defined as  $M=\mathbf{1}\left(\frac{1}{2 \sigma^2}>1\right)$, where $\mathbf{1}$ is the indicator function.

\subsubsection{Uncertainty-guided Refinement} Inspired by~\cite{yu2024viewcrafter}, we address the inconsistencies highlighted by the uncertainty map by applying a video denoising diffusion model~\cite{xing2025dynamicrafter} to the rendered sequence. This model leverages the masked areas identified earlier and restores the temporal and spatial consistency by ``filling in'' these regions with content that aligns seamlessly with the surrounding pixels. The diffusion model operates iteratively, refining each frame while considering the neighboring frames to ensure smooth transitions and maintain consistent visual quality. This step is crucial for preserving the flow of the sequence, reducing issues like jitter or flicker that can disrupt the viewer's experience. Once the sequence is refined and consistent, the updated frames are used to improve the 4D Gaussian field. This creates a feedback loop, aligning the 4D representation with the enhanced image sequence, which boosts the overall quality and stability of the 4D scene. Note that we condition the video diffusion model on the first and last frames of the input sequence to address the hallucination problem.

\subsection{Generalizable 3D Gaussian Field Predictor Learning}
\label{sec:unet-finetuning}
Aside from the inconsistency issues we've already improved, we also find that the quality of the 4D Gaussian fields doesn’t always match the improvements made by the image enhancer. This is expected as there is a notable domain gap between the pre-trained distribution of the image enhancer model~\cite{wang2018esrgan}, which is trained on DIV2K dataset~\cite{agustsson2017ntire}, and LGM~\cite{tang2025lgm}, which is trained on Objaverse~\cite{deitke2023objaverse}. To address this issue, we propose to fine-tune the U-Net model derived from LGM with the pre-processed Objaverse dataset. Specifically, we first follow LGM~\cite{tang2025lgm} to filter low-quality 3D models. In each training step, we randomly choose an input image with an elevation angle between -30 and 30 degrees. The MV-Adapter~\cite{huang2024mv} is then used to generate four orthogonal views, including the original image. These views are processed through the image enhancer and consistency refinement steps, and are subsequently passed into the U-Net model to produce the 4D Gaussian field. Finally, we render images from the Gaussian field based on the angles of the four orthogonal views for supervision. This training process allows the fine-tuned U-Net to reduce the domain gap between the pre-trained U-Net from LGM and the image enhancer model, resulting in improved quality of the 4D Gaussian fields.

\section{Experiments}
For the experiments, we conduct both qualitative and quantitative comparisons on 4D generation, perform ablation studies, and explore various applications of our method. 

\subsection{Implementation Details}
For the evaluation of video-to-4D generation, we utilize the video dataset provided by Consistent4D~\cite{jiangconsistent4d}. We employ the Segment Anything Model (SAM)~\cite{kirillov2023segment} to preprocess the input image sequences to extract the foreground objects. To evaluate image-to-4D generation, we curate a dataset by collecting images from the internet. These images are converted to RGBA format and resized to a resolution of $512\times512$ to ensure compatibility with our pipeline. For fine-tuning, we utilize the 80K 3D object subset\cite{tang2025lgm} of the Objaverse dataset~\cite{deitke2023objaverse} after filtering out low-quality models. Each 3D model is rendered into RGB images from 100 camera views at a resolution of $512\times512$.

The training process is being conducted using the asymmetric U-Net model on 4 NVIDIA V100 GPUs, with each GPU processing a batch size of 4 under bfloat16 precision. For each batch, a single camera view is being randomly sampled, while 4 orthogonal views are being generated using the MV-adapter~\cite{huang2024mv} based on the input view. The asymmetric U-Net model is generating the 3D Gaussian field, which is then being rendered into images for the orthogonal views. Original Objaverse 3D object rendered images are being used as supervision signals. The rendered 3D Gaussians are being compared to the original at a resolution of 512×512 using the mean squared error (MSE) loss. To optimize memory usage, images are being resized to 256×256 for LPIPS loss calculation. The AdamW optimizer is being employed with a learning rate of $4 \times 10^{-4}$, a weight decay of 0.05, and momentum parameters of $0.9$. The learning rate is following a cosine annealing schedule to gradually decay to zero during training. Gradients are being clipped to a maximum norm of 1.0 to enhance stability. Additionally, grid distortion and camera jitter are being applied with a probability of $50\%$ to improve generalization.

\begin{figure*}[t]
\centering
\includegraphics[width=15.5cm]{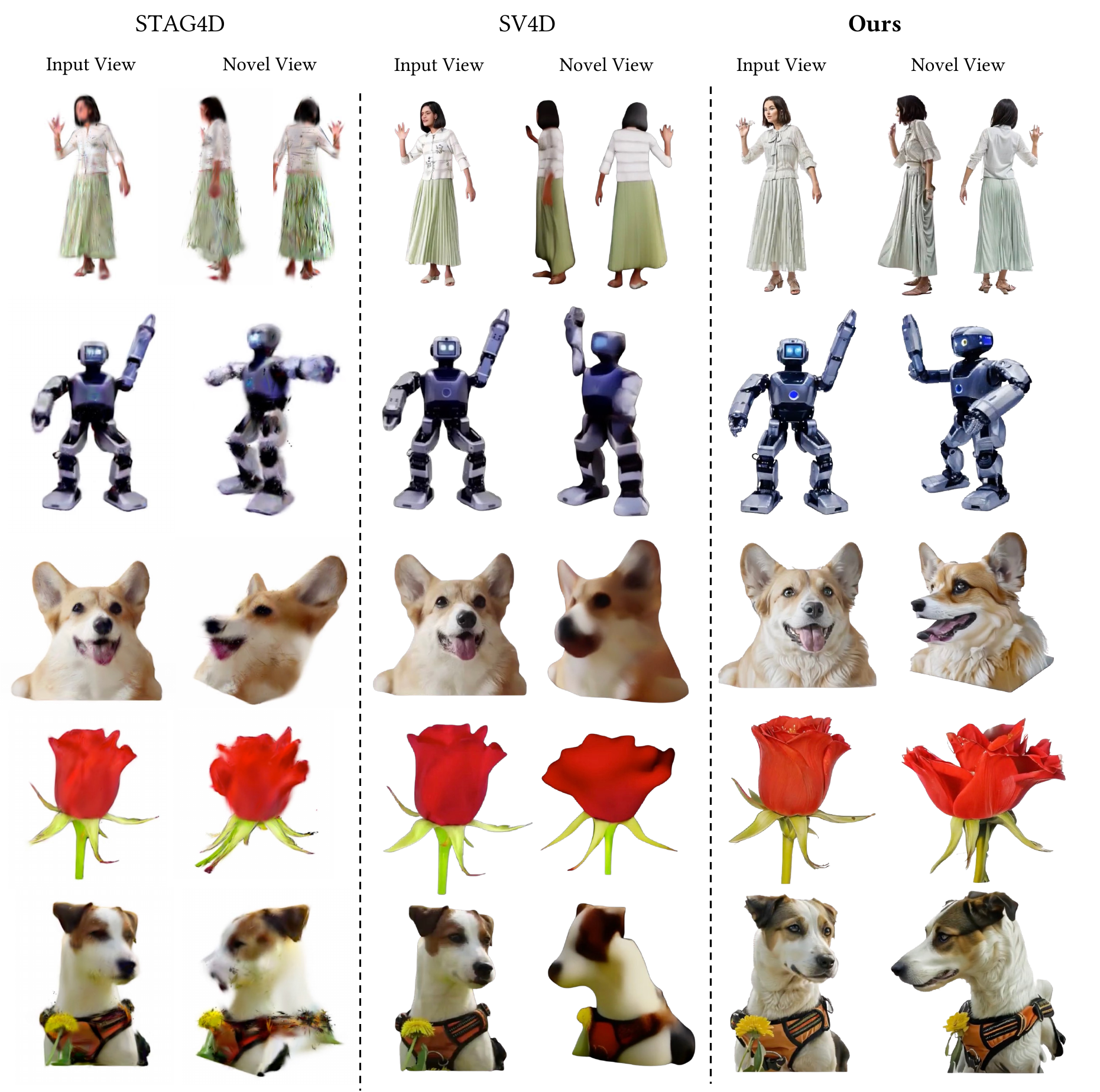}
\caption{\textbf{Comparison on \emph{video-to-4D} generation}. The rendered image of the input view from the 4D object is on the left column, and the rendered images of the novel view are illustrated on the two right columns. }
\centering
\label{fig:comparison}
\end{figure*}

\subsection{Main Comparison}

We compare our model with state-of-the-art baselines including STAG4D~\cite{zeng2024stag4d}, SV4D~\cite{xie2024sv4d},  Consistent4D~\cite{jiangconsistent4d}, 4Diffusion~\cite{zhang20244diffusion} and Diffusion4d~\cite{liang2024diffusiond}. As shown in~\figref{fig:comparison}, both STAG4D and SV4D suffer from producing satisfactory results when rendering novel views. This distinction is particularly pronounced in scenarios involving complex human structures, such as detailed facial features and loose clothing (first row). 
% While other methods often fail to reconstruct these intricate details accurately in both input and novel views, our model excels in preserving and reconstructing high-quality human features. 
These results underscore the superior capability of our method in performing robust video-to-4D reconstruction.

For \emph{video-to-4D} quantitative evaluation in~\tabref{tab:quali}, we assess the quality of each generated image by comparing it with its corresponding ground truth using metrics such as Learned Perceptual Similarity (LPIPS) and CLIP-Score (CLIP-S). These metrics help evaluate the visual fidelity and semantic alignment of the generated images. To measure temporal and spatial coherence in the generated video, we report the Fréchet Video Distance (FVD), a widely used video-level metric in video generation tasks. Our method achieves the best performance compared to all baselines across all evaluation metrics on Consistent4D dataset~\cite{jiangconsistent4d} and ObjaverseDy test set~\cite{deitke2023objaverse, xie2024sv4d} as shown in~\tabref{tab:quali}, indicating that our method is able to generate temporally and spatially coherent 4D content. For \emph{image-to-4D} evaluation in~\tabref{tab:quali_img}, we compare the generation results from measurements as in~\cite{liang2024diffusiond}. Our method still outperforms all baseline methods. %Specifically, FVD-F calculates the FVD across frames for each individual view, while FVD-V computes the FVD across different views at each frame. 
% These metrics comprehensively capture the visual quality and coherence of the generated 4D content.
% We surpass baseline methods in Learned Perceptual Similarity (LPIPS)~\cite{zhang2018unreasonable} and CLIP-Score (CLIP-S)~\cite{radford2021learning}, demonstrating higher visual fidelity and semantic consistency. 
% Additionally, our method achieves the lowest Fréchet Video Distance (FVD) values~\cite{unterthiner2018towards}, both at the frame level (FVD-F) and the view level (FVD-V), highlighting our model's capability to generate temporally and spatially coherent 4D content.

\begin{table}
  \caption{\textbf{\emph{Video-to-4D} quantitative Comparison on Consistent4D Dataset~\cite{jiangconsistent4d}.}}
  \label{tab:quali}
  \begin{tabular}{lcccc}
    \toprule
    Model&LPIPS$\downarrow$&CLIP-S$\uparrow$&FVD-F$\downarrow$&FVD-V$\downarrow$\\
    \midrule
    Consistent4D & 0.134&0.87 & 1133.93&735.79\\
    STAG4D & 0.126& 0.91 &992.21&685.23\\
    SV4D & 0.118 & 0.92 & 732.40&503.51\\
    4Diffusion & 0.13 & 0.94 & 489.2 & 405.5\\
    \textbf{Ours} & \textbf{0.090}& \textbf{0.97}& \textbf{390.85}&\textbf{282.79}\\
  \bottomrule
\end{tabular}
\end{table}

\begin{table}
  \caption{\textbf{\emph{Video-to-4D} Quantitative Comparison on ObjaverseDy Test Set~\cite{deitke2023objaverse, xie2024sv4d}.}}
  \label{tab:quali_ob}
%  \begin{adjustbox}{width=\linewidth}
  \begin{tabular}{lcccc}
    \toprule
    Model&LPIPS$\downarrow$&CLIP-S$\uparrow$&FVD-F$\downarrow$&FVD-V$\downarrow$\\
    \midrule
    Consistent4D & 0.165&0.896 & 880.54&488.38\\
    STAG4D & 0.158& 0.860 &929.10&453.62\\
    SV4D & 0.131 & 0.905 & 659.66&368.53\\
    \textbf{Ours} & \textbf{0.112}& \textbf{0.939}& \textbf{383.71}&\textbf{267.94}\\
  \bottomrule
\end{tabular}
%\end{adjustbox}
\end{table}

\begin{table}
  \caption{\textbf{Quantitative Comparison on \emph{Image-to-4D} Generation.}}
  \label{tab:quali_img}
  \begin{tabular}{lcccc}
    \toprule
    Model&LPIPS$\downarrow$&CLIP-S$\uparrow$&PSNR$\downarrow$&FVD$\downarrow$\\
    \midrule
    4DGen & 0.28&0.84 & 14.4&736.6\\
    STAG4D & 0.24& 0.86 &15.2&675.4\\
    Diffusion4D & 0.18 & 0.89 & 16.8&490.2\\
    \textbf{Ours} & \textbf{0.12}& \textbf{0.94}& \textbf{19.2}&\textbf{395.0}\\
  \bottomrule
\end{tabular}
\end{table}

\subsection{Ablation Study}
We conduct an ablation study on the Consistent4D dataset~\cite{jiangconsistent4d} and ObjaverseDy test set~\cite{deitke2023objaverse, xie2024sv4d}, to assess the impact of key components in our method. Specifically, we examine three variants:
no uncertainty masking, 2) no U-Net training, and 3) our full model.
The evaluation metrics include LPIPS, CLIP-S, FVD-F, and FVD-V, which respectively measure perceptual similarity, alignment with textual semantics, and spatial-temporal consistency.

% The experimental results are presented across three configurations in~\tabref{tab:ablation}: (1) results without uncertainty map masking, (2) results without asymmetric U-Net training, and (3) results using our full method. 
As shown in~\tabref{tab:ablation}, omitting either the uncertainty masking or U-Net training \emph{significantly degrades} LPIPS, CLIP-S and FVD metrics, demonstrating the importance of uncertainty map both components in handling spatial-temporal inconsistencies and preserving high-fidelity. In Supplementary Material, we present additional comparisons of model design choices, including the use of feedback loop, the incorporation of the image enhancer, and video diffusion model conditioning on first/last frames as discussed in~\secref{sec:refinement}.
% This ablation study underscores the synergistic role of these components in achieving state-of-the-art results for dynamic 4D content generation.

\begin{table}
  \caption{\textbf{Ablation Study.} The experiments are conducted on Consistent4D dataset~\cite{jiangconsistent4d}.}
  \label{tab:ablation}
  \begin{tabular}{lcccc}
    \toprule
    Model&LPIPS$\downarrow$&CLIP-S$\uparrow$&FVD-F$\downarrow$&FVD-V$\downarrow$\\
    \midrule
    w/o mask & 0.114&0.93 & 507.15&413.79\\
    w/o train & 0.107& 0.96 &445.33&364.81\\ 
    \textbf{Ours} & \textbf{0.090}& \textbf{0.98}& \textbf{390.85}&\textbf{282.79}\\
  \bottomrule
\end{tabular}
\end{table}

\subsection{\OMO for Different Applications}
Besides taking a single-view image or a text prompt as input to obtain 4D dynamic scenes, our method can also be applied to different tasks, including 4D human motion transfer, and text-guided 4D content editing.

\begin{figure*}[t]
\centering
\includegraphics[width=18cm]{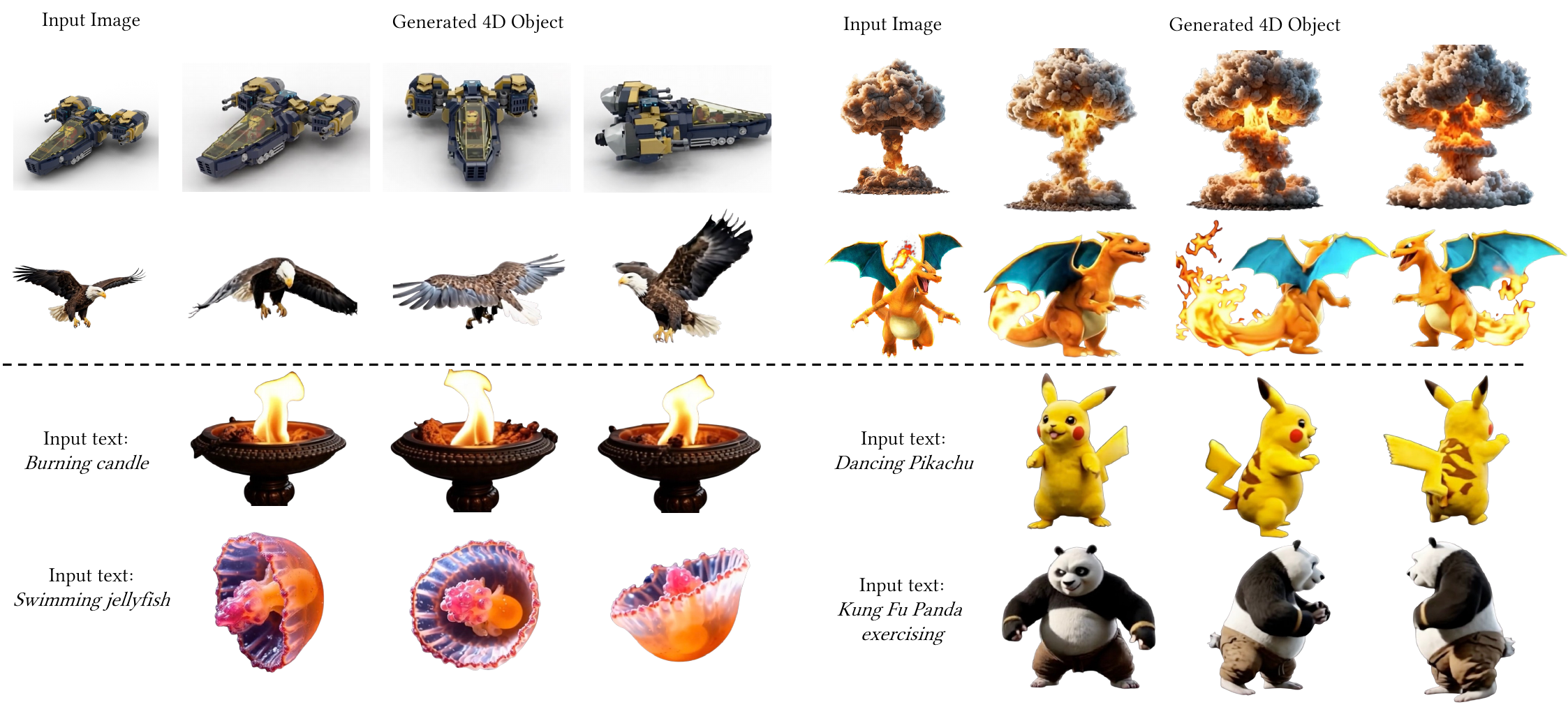}
\caption{\textbf{4D Content Creation with Text/Image as Input}. The first two rows are results with image inputs, and bottom two rows are results with text inputs.}
\centering
\label{fig:textandimage}
\end{figure*}

\begin{figure}[hbp]
\includegraphics[width=\linewidth]{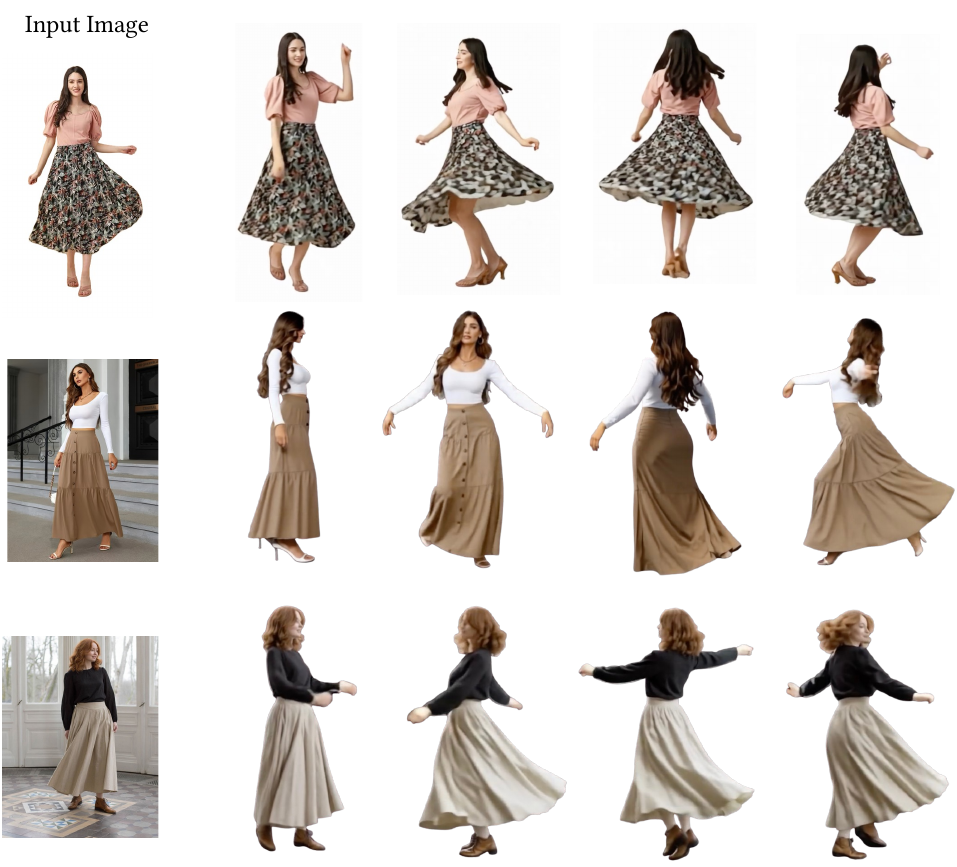}
\caption{\textbf{4D Human Generation with an Input Image as Guidance}. The first row shows the input image, while the subsequent rows depict rendered novel views under various poses.}
\label{fig:human}
\end{figure}
\begin{figure}
%\centering
\includegraphics[width=\linewidth]{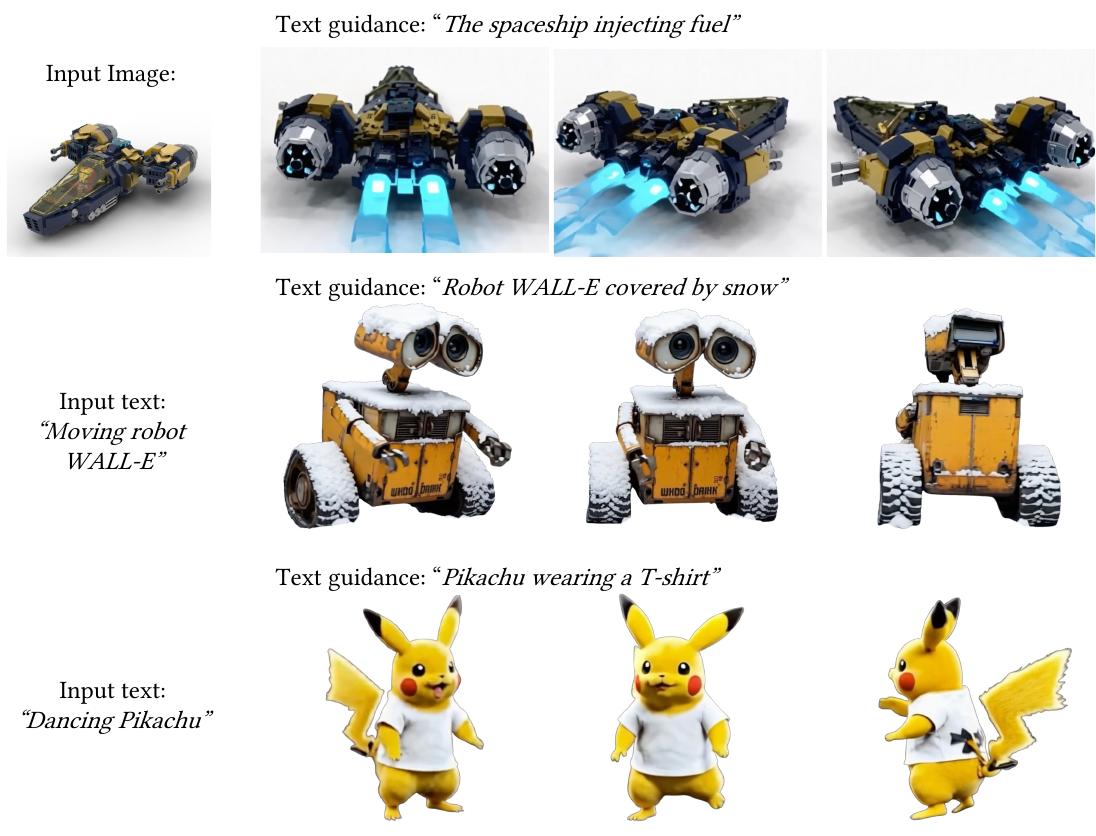}
\caption{\textbf{4D Content Editing with Text Guidance}. The first column showcases the original input (text or image), while the subsequent three columns present the edited outputs. Each edited 4D object is displayed beneath the corresponding text.}
%\centering
\label{fig:edit}
\end{figure}

\subsubsection{Text/Image Conditioned 4D Generation} To demonstrate our method’s capability for \emph{text-to-4D} and \emph{image-to-4D} generation, we show our generation results in ~\figref{fig:textandimage}. For \emph{text-to-4D generation}, we first employ a text-to-image diffusion model to convert the input textual prompt into a high-quality image and then combine that image with the original text in a stable video diffusion model to produce a coherent short video. In \emph{image-to-4D generation}, the pipeline simply begins with an input image, bypassing the text-to-image step. From the resulting videos, we use a multi-view diffusion model to generate four orthogonal view sequences, which are then fed into our reconstruction pipeline to construct a 4D Gaussian field. 

As shown in \figref{fig:textandimage}, the top two rows illustrate \emph{image-to-4D} results, while the bottom two rows depict \emph{text-to-4D} outputs. The generated 4D reconstructions demonstrate the effectiveness of our pipeline in maintaining structural coherence and high visual fidelity across multiple perspectives. For the \emph{image-to-4D} examples, we observe precise alignment and consistent detail retention in novel views derived from the input image. As for the \emph{text-to-4D} results, the generated scenes accurately align with the semantic content of the input text, producing dynamic and visually plausible outputs.

\subsubsection{4D Human Generation} Given a source video with the desired motion to be transferred and an image of the human subject to be animated, we first use a pose extraction model~\cite{kocabas2020vibe} to detect key body landmarks, skeletal poses, and motion trajectories. Next, we apply Champ~\cite{zhu2024champ}, a 2D motion transfer model, to animate the input image, making it move according to the extracted motion. Our method then uses the resulting animated image sequence $\{I_t | t \in [1, T]\}$, where $T$ is the total number of frames, to generate the corresponding 4D Gaussian scenes.
\figref{fig:human} showcases the results of our \emph{4D human generation} pipeline, which combines a single input image with a motion sequence to produce high-fidelity, dynamic human representations. First, the input image and motion sequence are processed through a 2D motion transfer model to create a video of the subject performing the specified action (see details in Supplementary Material). Next, we follow our pipeline and apply a multi-view diffusion model and construct 4D Gaussians of the human.

\figref{fig:human} illustrates the effectiveness of our approach. These results highlight our ability to preserve intricate human details, including complex structures like facial details and loose clothing, across varying views and motions. 

\subsubsection{Text-guided Editing} For this application, we use the InstructPix2Pix~\cite{brooks2023instructpix2pix} network to modify the output video generated by the video diffusion model, guided by a text prompt (see Fig.~\ref{fig:pipeline}). For instance, starting with a video of a house, the pix2pix network is applied to transform the video based on a prompt like ``house on fire''. Specifically, the pix2pix network performs image-to-image translation, adjusting each frame of the video to match the specified scene changes. Once the transformation is complete, the modified video sequence is used to refine the corresponding 4D Gaussian sequence, resulting in a final 4D content that accurately reflects the updated dynamics of the ``house on fire'' scenario.
In~\figref{fig:edit}, we showcase our method's \emph{text-guided 4D editing} capability that transforms 4D Gaussian representations based on user-specified textual or visual prompts. Starting from our 4D Gaussian field, we employ a pix2pix network to edit the rendered video according to the guidance text, producing an updated video sequence. This sequence is further optimized using the 4D Gaussian representation, ensuring coherence and alignment with the guidance.

The results are illustrated in~\figref{fig:edit}.
These examples demonstrate the effectiveness of our approach in introducing realistic and coherent transformations, such as attribute changes or new dynamic effects, while maintaining high fidelity to the original 4D content structure.

\section{Conclusions}

In this paper, we introduce a novel framework for high-quality 4D content generation, which addresses key challenges in dynamic scene creation by leveraging a 4D Gaussian splatting representation. Our method demonstrates strong generalization capabilities, enabling the generation of temporally stable and high-fidelity 4D content from monocular videos, images, and text prompts. Through careful integration of a multi-view video diffusion model and an asymmetric U-Net, we improve both spatial and temporal consistency, enhancing the visual coherence of the generated scenes. Our ablation studies validate the importance of components such as uncertainty map masking and asymmetry U-Net training for improving the quality of 4D content generation. The proposed framework is versatile and can be applied to a variety of scenarios, including text/image conditioned 4D generation, 4D human generation, and text-guided content editing. We believe that our approach marks a significant advancement in 4D scene generation, offering a robust solution that balances computational efficiency with high-quality results for real-world applications in digital humans, gaming, AR/VR, and media production.

\begin{acks}
The work is supported by the Hong Kong Research Grants Council - General Research Fund (Grant No.: 17211024). 
\end{acks}
% \section{Final copy}

% You must include your signed IEEE copyright release form when you submit your finished paper.
% We MUST have this form before your paper can be published in the proceedings.

% Please direct any questions to the production editor in charge of these proceedings at the IEEE Computer Society Press:
% \url{https://www.computer.org/about/contact}.
%{
%    \small
%    \bibliographystyle{ieeenat_fullname}
%    \bibliography{main}
%}

%\bibliographystyle{ACM-Reference-Format}
%\bibliography{Splat4D}

\bibliographystyle{ACM-Reference-Format}
\bibliography{Splat4D}

%\title{Splat4D: Diffusion-Enhanced 4D Gaussian Splatting for Temporally and Spatially Consistent Content Creation\\--Supplementary Material--}
  
%\maketitle
\twocolumn[

  {\Huge Splat4D: Diffusion-Enhanced 4D Gaussian Splatting for Temporally and Spatially Consistent Content Creation\\--\textit{Supplementary Material}--\\[1em]}
%  {\large Minghao Yin}\\
%  The University of Hong Kong\\[2em]
%  \today

\vspace{1em}
]

%\begin{titlepage}
%{\Huge Splat4D: Diffusion-Enhanced 4D Gaussian Splatting for Temporally and Spatially Consistent Content Creation\\--Supplementary Material-- \par}
%\end{titlepage}
%\begin{center}
%{Splat4D: Diffusion-Enhanced 4D Gaussian Splatting for Temporally and Spatially Consistent Content Creation\\--Supplementary Material--}
%\end{center}

%\clearpage
%\setcounter{page}{1}
%\maketitlesupplementary

%\newpage
%%
%% If your work has an appendix, this is the place to put it
\appendix

\section{Details for 4D Human Generation}
The process begins with a source video containing the desired motion for transfer. A pose extraction model~\cite{kocabas2020vibe} is used to capture SMPL~\cite{loper2023smpl} sequences, key body landmarks, skeletal poses, and motion trajectories over time. This model processes each frame to extract temporal motion data while preserving dynamic details such as joint rotations, limb movements, and fine-grained motion nuances.

Once the SMPL motion data is obtained, it is projected onto four orthogonal views to serve as input for multi-view generation via the MV-Adapter~\cite{huang2024mv}. The projections include sequences of 2D depth images, normal maps, human joint images, and semantic segmentation images. These multi-view motion representations are then fed into the Champ~\cite{zhu2024champ} 2D motion transfer model, which generates four sequences of motion images. This model takes static images of the target subject as input and, guided by the extracted pose data, produces motion sequences that replicate the source motion while preserving the visual identity and appearance of the target subject. These sequences serve as an intermediate representation, bridging the source motion and the final 4D output.

The generated 2D motion sequences, denoted as $\{I_t | t \in [1, T]\}$, are subsequently used to initialize a 4D Gaussian field in our framework. Each frame at time $t$ is represented as a set of Gaussians $\mathcal{G}(\mathcal{S}, t)=\left[\mathcal{X}_t, s_t, r_t, \sigma_t, \zeta_t\right]$. Using our U-Net-based refinement and Gaussian splatting pipeline, the initial field is optimized to ensure temporal and spatial consistency across views. This process captures intricate details such as loose clothing and rapid movements, resulting in a coherent and realistic 4D human representation.

\section{Additional Ablation Study}
\begin{figure}
\centering
\includegraphics[width=\linewidth]{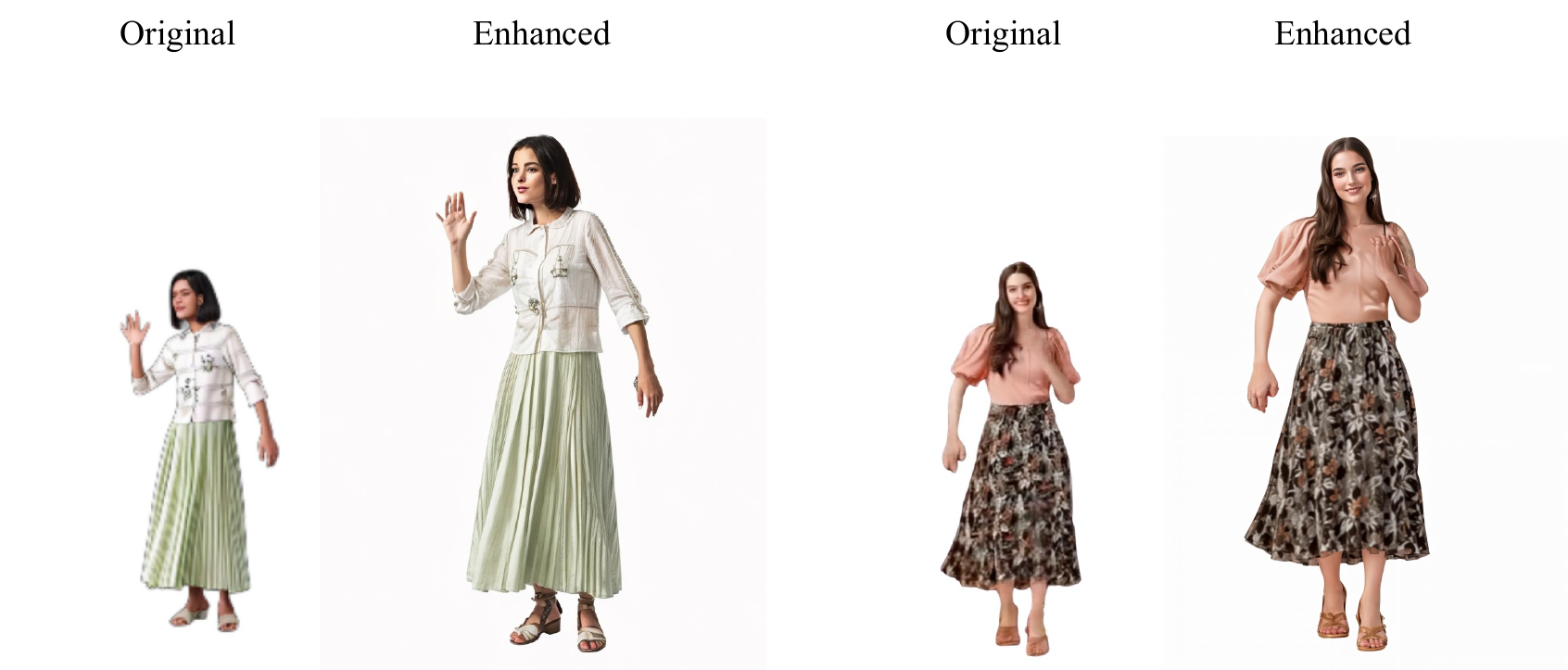}
\caption{\textbf{Effect of Image Enhancer.} We show the difference between human images before and after enhancement with the image enhancer. The enhanced images contain more fine details.}
\centering
\label{fig:enhancer}
\end{figure}

%\subsection{}
In~\tabref{tab:appendix_ablation}, we perform an ablation study focusing on multiple factors: the inclusion of the feedback loop, the image enhancer, the video diffusion model condition, the use of MV-Adapter compared to SV4D~\cite{xie2024sv4d}, as detailed in~Section 3.2. The experiments are conducted on Consistent4D~\cite{jiangconsistent4d}. Results demonstrate that using feedback loop optimization, the image enhancer improves visual quality. Condition video diffusion model with first and last frames of input sequence can enhance generation performance. Furthermore, replacing SV4D with MV-Adapter leads to better reconstruction results. These findings highlight the importance of both components in achieving high-quality dynamic scene generation. As discussed in~Section 3.2, we illustrate the images before and after the image enhancer process in ~\figref{fig:enhancer}. 

\begin{table}[htbp]
  \caption{\textbf{Additional Ablation Study.} The first row illustrates the results without image enhancer. The second row shows the results using SV4D over MV-Adapter for multi-view generation.}
  \label{tab:appendix_ablation}
  \begin{tabular}{lcccc}
    \toprule
    Model&LPIPS$\downarrow$&CLIP-S$\uparrow$&FVD-F$\downarrow$&FVD-V$\downarrow$\\
    \midrule
    w/o loop & 0.120 & 0.90 & 831.84 &473.91 \\
    w/o enhancer & 0.108&0.96 & 463.39&384.26\\ 
    ${SV4D}^*$ & 0.105& 0.94 &441.72&356.25\\
    w/o cond & 0.101 & 0.94 & 425.72 & 339.05 \\ 
    \textbf{Ours} & \textbf{0.090}& \textbf{0.98}& \textbf{390.85}&\textbf{282.79}\\
  \bottomrule
\end{tabular}
\end{table}

\begin{table}[htbp]
  \caption{\textbf{Evaluation on Liquid Case.}}
  \label{tab:appendix_liquid}
  \begin{tabular}{lccc}
    \toprule
    Model&CLIP$\uparrow$&LPIPS$\downarrow$&FVD$\downarrow$\\
    \midrule
    Consistent4D ~\cite{jiangconsistent4d} & 0.78&0.166& 1282.5\\
    STAG4D~\cite{zeng2024stag4d} & 0.80& 0.160 &1231.9\\
    SV4D~\cite{xie2024sv4d} & 0.88 & 0.147 & 772.6 \\
    \textbf{Ours} & \textbf{0.93}& \textbf{0.127}& \textbf{493.0}\\
  \bottomrule
\end{tabular}
\end{table}

\begin{table}[htbp]
  \caption{\textbf{Evaluation on Multi-object Case.}}
  \label{tab:appendix_multiobj}
  \begin{tabular}{lccc}
    \toprule
    Model&CLIP$\uparrow$&LPIPS$\downarrow$&FVD$\downarrow$\\
    \midrule
    Consistent4D~\cite{jiangconsistent4d} & 0.82&0.138& 1190.7\\
    STAG4D~\cite{zeng2024stag4d} & 0.85& 0.144 &1028.2\\
    SV4D~\cite{xie2024sv4d} & 0.90 & 0.125 & 722.3 \\
    \textbf{Ours} & \textbf{0.96}& \textbf{0.102}& \textbf{428.5}\\
  \bottomrule
\end{tabular}
\end{table}

\begin{figure*}
\centering
\includegraphics[width=0.85\textwidth]{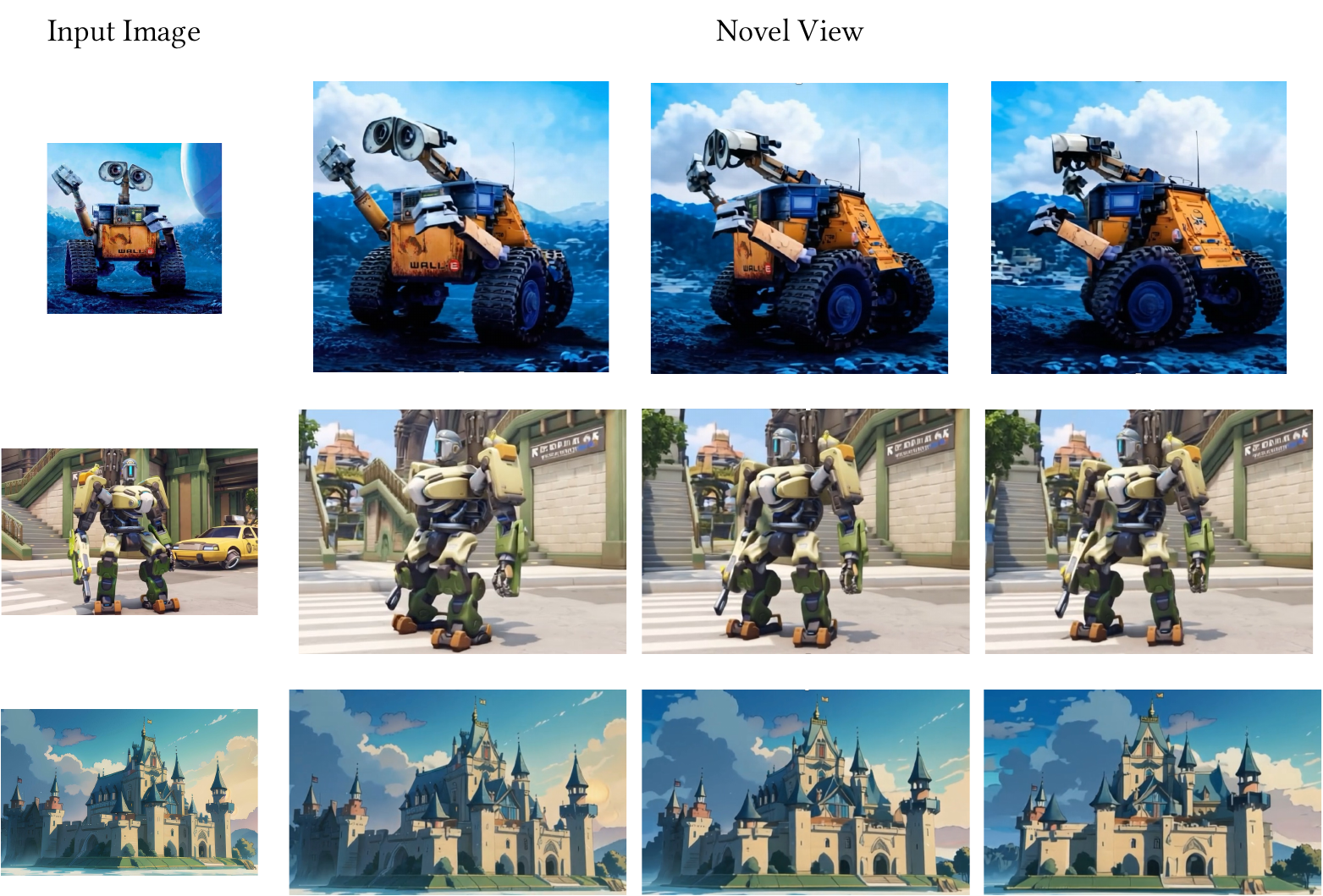}
\caption{\textbf{4D Generation Results with Background.} The first column on the left shows the input images, while the following three columns show the generation results with a bounded background.}
\centering
\label{fig:background}
\end{figure*}

\section{Test on Complex Scenarios}

In Fig.5 of the main paper, our method successfully handles dancing humans with non-rigid loose clothing. In order to measure the performance on more complex domains, we test on fluids (``splashing water'' from Consistent4D~\cite{jiangconsistent4d}) and multi-object interactions (``bouncing ball'' from DNeRF~\cite{pumarola2021d}, front-view video). The quantitative results for the liquid and multi-object scenario are reported below. Experimental results show that our model can handle complex scenarios such as cloth, fluids and multi-object interactions.

\section{More Visualizations}
\begin{figure*}[htbp]
\centering
\includegraphics[width=\textwidth]{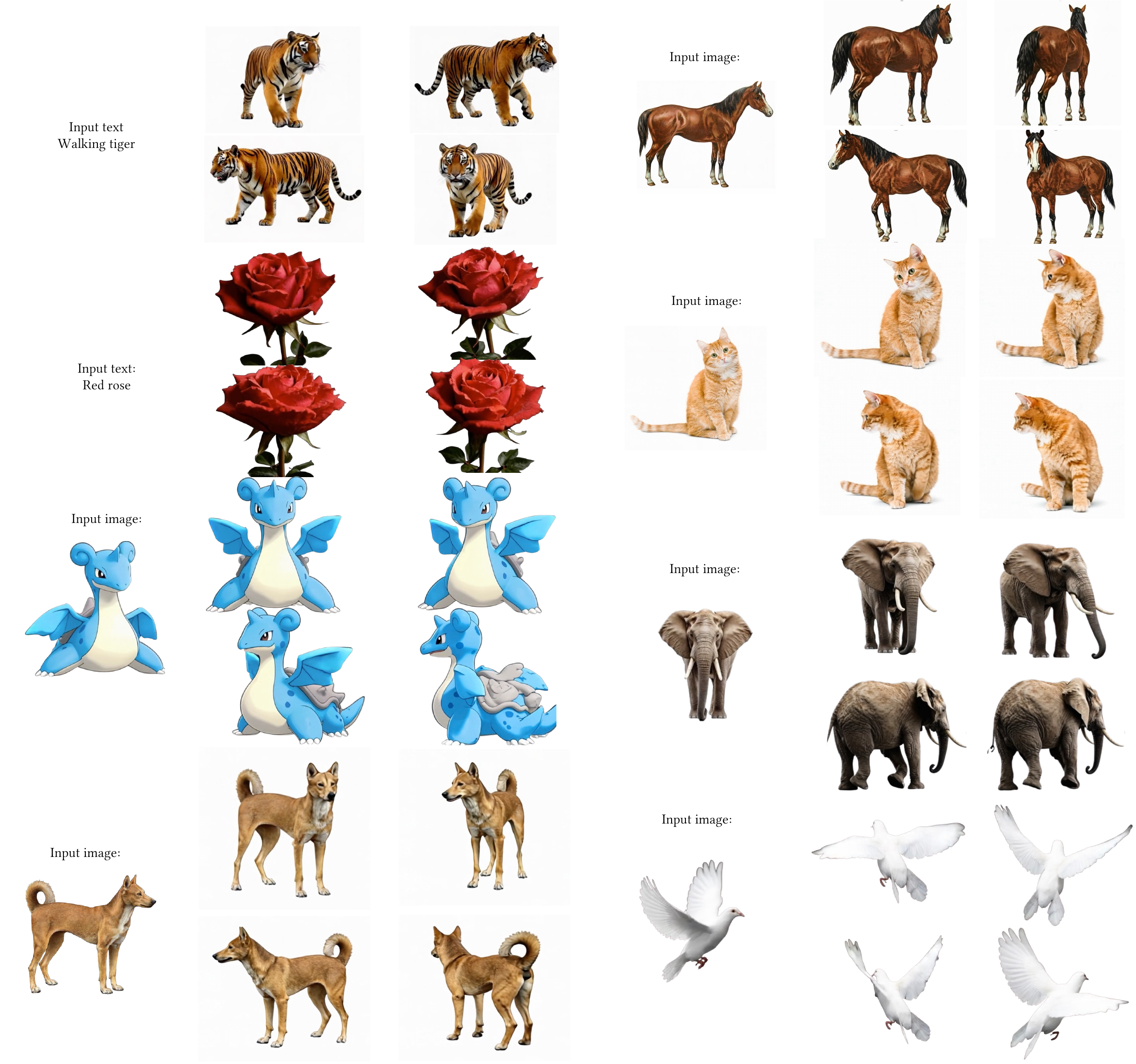}
\caption{\textbf{4D Generation Results Conditioned on Text/Image.} We present additional text/image-to-4D generation results. On the left are the inputs, and on the right are the generated 4D objects.}
\label{fig:app2}
\end{figure*}

\begin{figure*}[htbp]
%\centering
\includegraphics[width=16cm]{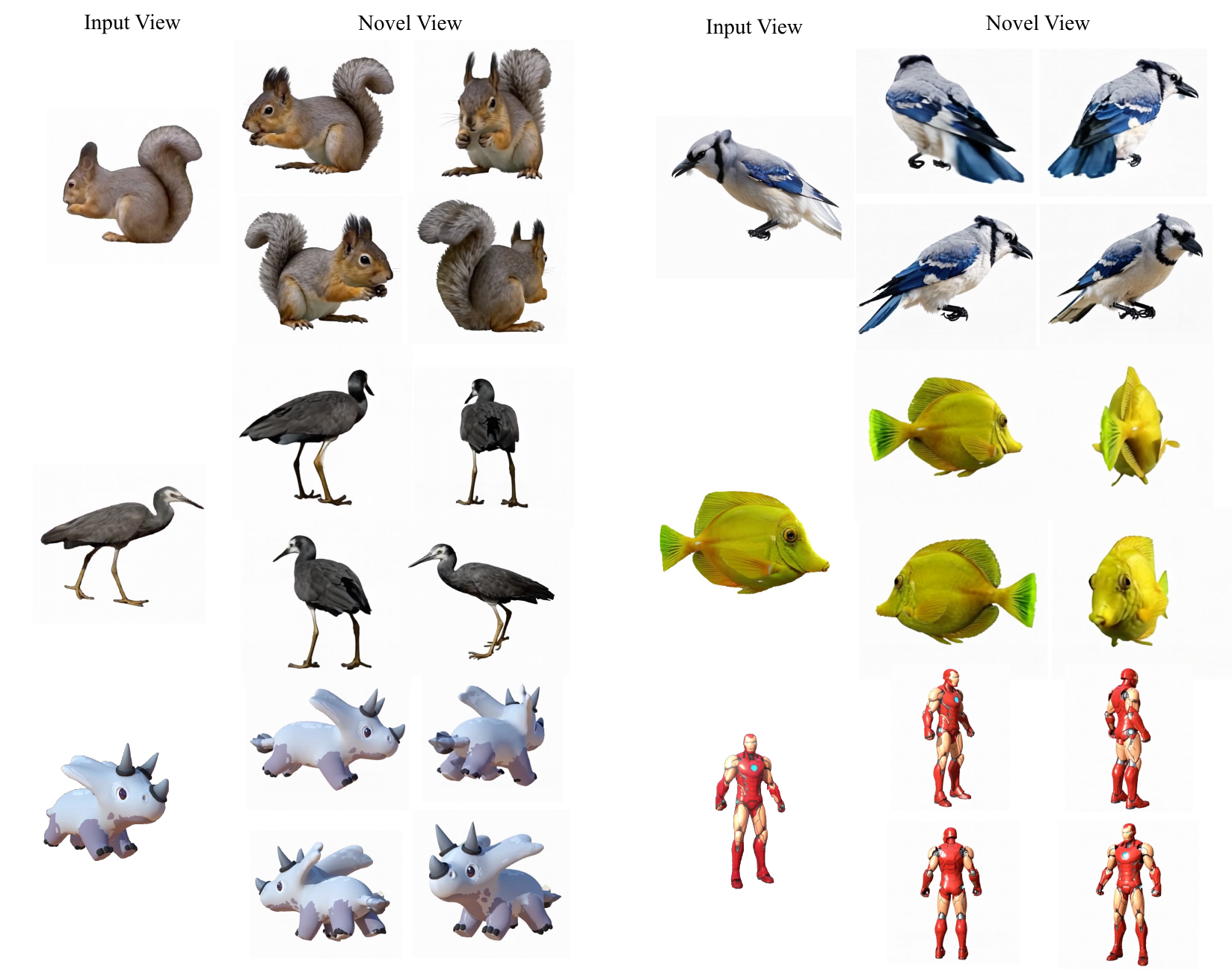}
\caption{\textbf{Video-to-4D Generation}. The left column shows the rendered image of the input view for the 4D object. The right columns show rendered images of novel views.}
%\centering
\label{fig:appendix}
\end{figure*}

In~\figref{fig:enhancer}, we show the difference between original image enhanced image. In~\figref{fig:background} we show our model can handle more challenging cases with bounded 4D scene. In~\figref{fig:app2} and~\figref{fig:appendix} we show more text/image to 4D generation results.

%\begin{figure}[htbp]
%\centering
%\includegraphics[width=8.2cm]{figs/appendix2.pdf}
%\caption{\textbf{4D Generation Results Conditioned on Text/Image.}}
%\label{fig:app2}
%\end{figure}

%\begin{figure}[htbp]
%\centering
%\includegraphics[width=7.5cm]{figs/imageto4d.pdf}
%\caption{\textbf{4D Generation Results Conditioned on Image.}}
%\label{fig:app3}
%\end{figure}

%\bibliographystyle{ACM-Reference-Format}
%\bibliography{Splat4D}

\end{document}